\documentclass[conference]{IEEEtran}
\IEEEoverridecommandlockouts
\usepackage{cite}
\usepackage{amsmath,amssymb,amsfonts}
\usepackage{algorithmic}
\usepackage{graphicx}
\usepackage{textcomp}
\usepackage{xcolor}
\usepackage{booktabs}    
\usepackage{multirow}     
\usepackage{graphicx}     
\usepackage{array}        
\def\BibTeX{{\rm B\kern-.05em{\sc i\kern-.025em b}\kern-.08em
    T\kern-.1667em\lower.7ex\hbox{E}\kern-.125emX}}
\begin{document}

\title{Multimodal Slice Interaction Network Enhanced by Transfer Learning for Precise Segmentation of Internal Gross Tumor Volume in Lung Cancer PET/CT Imaging\\

\thanks{This work was supported by the National Cancer Institute of the National Institutes of Health (award number 
R25CA288263 and R37CA229417).}
}

\author{
\IEEEauthorblockN{1\textsuperscript{st} Yi Luo}
\IEEEauthorblockA{
    \textit{Dept. of Biomedical Engineering} \\
    \textit{Johns Hopkins University}\\
    Baltimore, USA \\
    yluo62@jh.edu
}
\and
\IEEEauthorblockN{2\textsuperscript{nd} Yike Guo}
\IEEEauthorblockA{
    \textit{Dept. of Biomedical Engineering} \\
    \textit{Johns Hopkins University}\\
    Baltimore, USA \\
    yguo122@jh.edu
}
\and
\IEEEauthorblockN{3\textsuperscript{rd} Hamed Hooshangnejad}
\IEEEauthorblockA{
    \textit{Dept. of Radiation Oncology} \\
    \textit{and Molecular Radiation Sciences}\\
    \textit{Johns Hopkins University}\\
    Baltimore, USA \\
    h.hooshangnejad@gmail.com
}
\and
\IEEEauthorblockN{4\textsuperscript{th} Rui Zhang}
\IEEEauthorblockA{
    \textit{Dept. of Surgery} \\
    \textit{University of Minnesota}\\
    Minneapolis, USA \\
    ruizhang@umn.edu
}
\and
\IEEEauthorblockN{5\textsuperscript{th} Xue Feng}
\IEEEauthorblockA{
    \textit{Dept. of Biomedical Engineering} \\
    \textit{University of Virginia}\\
    Charlottesville, USA \\
    xf4j@virginia.edu
}
\and
\IEEEauthorblockN{6\textsuperscript{th} Quan Chen}
\IEEEauthorblockA{
    \textit{Comprehensive Cancer Center} \\
    \textit{Mayo Clinic}\\
    Phoenix, USA \\
    chen.quan@mayo.edu
}
\and
\IEEEauthorblockN{7\textsuperscript{th} Wil Ngwa}
\IEEEauthorblockA{
    \textit{Dept. of Radiation Oncology} \\
    \textit{and Molecular Radiation Sciences}\\
    \textit{Johns Hopkins University}\\
    Baltimore, USA \\
    wngwa1@jhmi.edu 
}
\and
\IEEEauthorblockN{8\textsuperscript{th} Kai Ding}
\IEEEauthorblockA{
    \textit{Dept. of Radiation Oncology} \\
    \textit{and Molecular Radiation Sciences}\\
    \textit{Johns Hopkins University}\\
    Baltimore, USA \\
    kai@jhu.edu 
}
}

\maketitle

\begin{abstract}
Lung cancer remains the leading cause of cancer-related deaths globally. Accurate delineation of internal gross tumor volume (IGTV) in PET/CT imaging is pivotal for optimal radiation therapy in mobile tumors such as lung cancer to account for tumor motion, yet is hindered by the limited availability of annotated IGTV datasets and attenuated PET signal intensity at tumor boundaries. In this study, we present a transfer learning-based methodology utilizing a multimodal interactive perception network with MAMBA, pre-trained on extensive gross tumor volume (GTV) datasets and subsequently fine-tuned on a private IGTV cohort. This cohort constitutes the PET/CT subset of the Lung-cancer Unified Cross-modal Imaging Dataset (LUCID). To further address the challenge of weak PET intensities in IGTV peripheral slices, we introduce a slice interaction module (SIM) within a 2.5D segmentation framework to effectively model inter-slice relationships. Our proposed module integrates channel and spatial attention branches with depthwise convolutions, enabling more robust learning of slice-to-slice dependencies and thereby improving overall segmentation performance. A comprehensive experimental evaluation demonstrates that our approach achieves a Dice of 0.609 on the private IGTV dataset, substantially surpassing the conventional baseline score of 0.385. This work highlights the potential of transfer learning, coupled with advanced multimodal techniques and a SIM to enhance the reliability and clinical relevance of IGTV segmentation for lung cancer radiation therapy planning.
\end{abstract}


\section{Introduction}
Lung cancer is the leading cause of cancer-related mortality worldwide\cite{leiter2023global}, with more than 238,000 new diagnoses annually in the US\cite{siegel2023cancer}. The overall five-year survival rate for lung cancer remains below 20\%\cite{fitzmaurice2017global}, posing a significant public health burden, even with the considerable advances in medical research and treatment approaches. It is predominantly diagnosed as non-small-cell lung cancer (NSCLC)\cite{bodor2020biomarkers}, for which stereotactic body radiation therapy is often the preferred treatment method when it is inoperable and early-stage\cite{luo2025language, hooshangnejad2025quantitative, hooshangnejad2023deepperfect}.  

Positron Emission Tomography/Computed Tomography (PET/CT) serves as a pivotal modality in the evaluation of lung tumors, delivering critical metabolic and anatomical insight for clinical decision-making\cite{xiang2022modality}. A key step in radiotherapy planning is the delineation of tumor volumes. The gross tumor volume (GTV) corresponds to the visible extent of the tumor detected via imaging modalities, while the internal gross tumor volume (IGTV) is defined as an extension of GTV that accounts for tumor movement during respiration, such as lung cancer radiation therapy planning, thereby ensuring adequate treatment coverage\cite{wambersie1999icru}. As a result, accurate delineation of the IGTV is essential for effective radiation therapy. PET acquisition times are inherently long, typically on the order of 15-30 minutes and 2-5 minutes per bed position over numerous breathing cycles.  As a result, the tumor motion is recorded as blurring, reduced standardized uptake values (SUV) as image features. However, poor image quality, motion artifacts, complex and irregular tumor morphology, intrinsically low PET signal intensity at the margins of the lesion, and high uptake in adjacent organs on PET/CT scans often lead to indistinct boundaries, thus complicating reliable IGTV segmentation. In recent years, the integration of AI into cancer research and clinical protocols has initiated profound transformations within the field \cite{luo2025opportunities}. Deep learning models have contributed significantly to the advancement of medical image analysis, enabling automated lesion detection, segmentation, and classification with improved accuracy and efficiency \cite{litjens2017survey}. Leveraging large annotated datasets, these models can learn complex representations of tumor characteristics, outperforming traditional image processing methods.

Despite the success of deep learning models in medical image analysis, their performance is frequently constrained by a dependence on large well-annotated datasets. In the context of IGTV segmentation from lung tumor PET/CT scans, the acquisition of sufficient, high-quality labeled data is often impractical due to the complexity of annotation and the limited availability of expert clinicians. This data scarcity not only impedes robust model training, but also restricts the generalizability of learned representations to diverse clinical scenarios. Consequently, there is an urgent need for strategies that can alleviate the impact of limited annotations on model performance.

In contrast, the publicly available PCLT20k dataset comprises 21,930 PET/CT image pairs from 605 unique patients\cite{mei2025cross}, and is primarily dedicated to lung tumor GTV segmentation, providing a substantial resource for model development. To address the scarcity of annotated IGTV data and fully exploit the abundance of GTV-labeled samples, we adopt a transfer learning strategy. In this work, we curate an IGTV PET/CT cohort comprising 1,067 high-quality PET/CT image pairs from 60 patients at Johns Hopkins Hospital. This cohort constitutes the PET/CT subset of the Lung-cancer Unified Cross-modal Imaging Dataset (LUCID) and is hereafter referred to as LUCID-PET/CT. Our approach pre-trains the network on PCLT20k and then fine-tunes it on LUCID-PET/CT, enabling robust feature learning and improving IGTV delineation with sparse annotations.

To perform IGTV segmentation, we employ the state-of-the-art cross-modal interactive perception network (CIPA) with the Mamba architecture\cite{mei2025cross}. This model integrates PET and CT image features through channel-wise rectification module (CRM) and dynamic cross-modality interaction module (DCIM), enabling effective fusion of metabolic and anatomical information from both modalities for accurate lung tumor segmentation.

In previous work, GTV segmentation methods typically employed a 2D approach, where each input consists of a single PET image and its corresponding CT image. However, this 2D paradigm is insufficient for accurate IGTV segmentation ignoring spatial information between slices, particularly in peripheral areas with weak PET signal at the tumor boundaries, where distinguishing true margins becomes especially challenging.

\begin{figure*}[t]
\centering
\includegraphics[width=\textwidth]{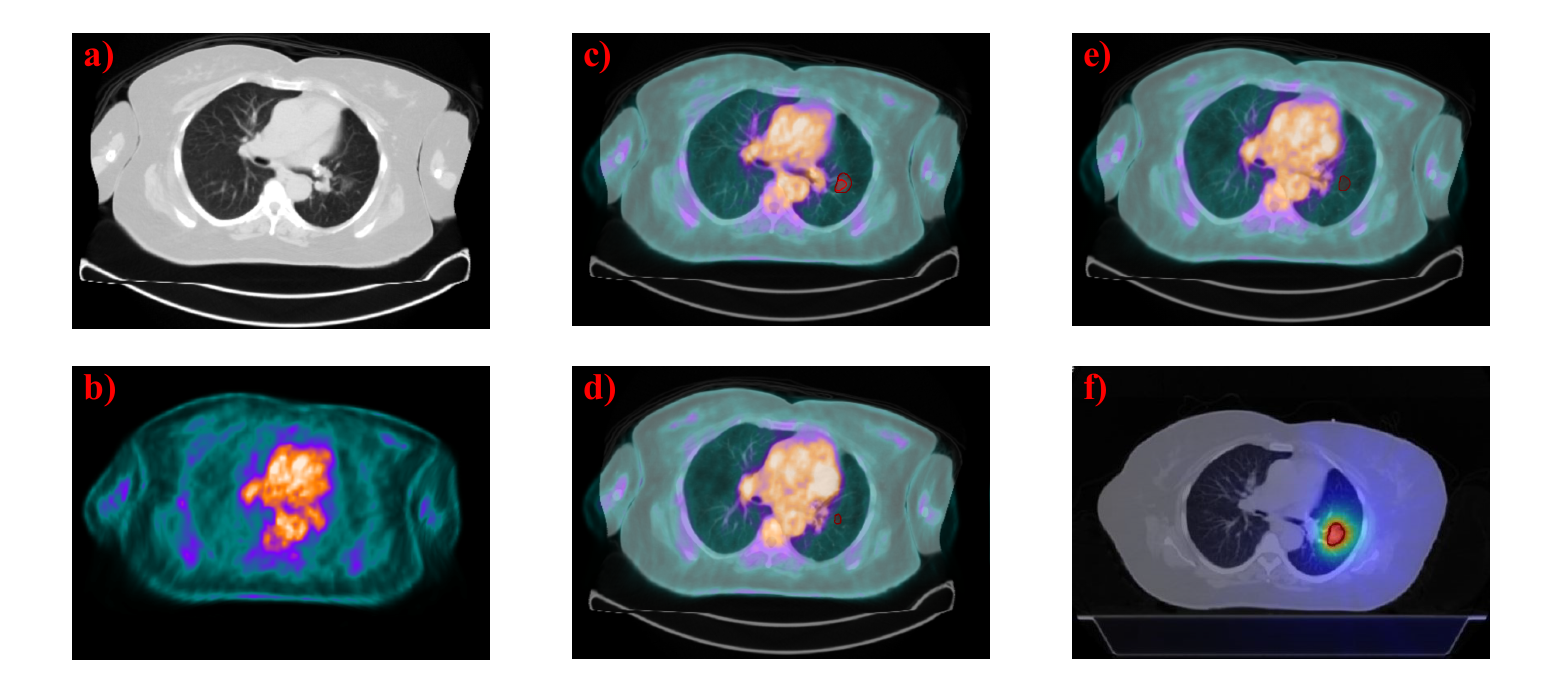}
\caption{Multimodal illustration of IGTV versus GTV and radiotherapy planning. (a) CT image; (b) PET image; (c) fused PET/CT image with IGTV (dark red) and GTV (light red) contours; (d,e) IGTV marginal slices with high organ SUV and low tumor uptake; (f) radiation therapy dose distribution derived from the IGTV contour.}
\label{fig:intro}
\end{figure*}

To concretize the challenges mentioned above and motivate our design choices, Figure~\ref{fig:intro}  presents a representative case. The CT (Fig. 1a) and PET (Fig. 1b) provide complementary anatomic and metabolic cues, while the fused image (Fig. 1c) highlights the discrepancy between the IGTV (dark red) and GTV (light red) contours. The marginal slices in Figs. 1d and 1e illustrate the confounding effect of high SUV uptake in neighboring organs (e.g., the heart) and low tumor uptake, which obscures lesion borders on PET and hampers robust delineation. Finally, Fig. 1f visualizes the radiation therapy dose distribution planned on the PTV, which is a 5 mm uniform expansion from IGTV to account for delivery uncertainty and patient setup error, underscoring the clinical impact of precise IGTV definition.

To address these limitations, we propose a 2.5D segmentation strategy that incorporates three consecutive slices as input to the transfer learning network, providing richer spatial context and enhancing the model’s ability to capture local anatomical continuity. Furthermore, to explicitly model inter-slice relationships and facilitate robust feature learning across adjacent slices, we introduce a Slice Interaction Module (SIM). This lightweight module consists of three complementary branches: a channel attention branch that adaptively weights the importance of each slice, a spatial attention branch that highlights salient regions across slices, and a slice relation branch employing depthwise convolutions to model complex inter-slice dependencies. By fusing the outputs from these branches, SIM generates refined feature representations that enable the network to more accurately distinguish ambiguous boundaries and improve segmentation performance, particularly in regions with weak PET signal intensity.

In summary, our contributions are as follows:
\begin{itemize}
    \item We propose a transfer learning framework for IGTV segmentation, leveraging a large-scale GTV labeled PET/CT dataset for pre-training and a private IGTV annotated dataset for fine-tuning.
    
    \item We introduce a 2.5D segmentation strategy incorporating a lightweight SIM that captures inter-slice relationships through channel attention, spatial attention, and slice relation modeling.
    
    \item Extensive experiments on both public and private datasets demonstrate that our approach achieves superior IGTV segmentation performance compared to conventional baselines, with a Dice of 0.609, which is a 58.2\% improvement over the baseline value of 0.385. Our framework provides a practical solution for IGTV contouring in radiotherapy planning.
\end{itemize}

\section{Related Work}

\subsection{Lung Tumor Segmentation in PET/CT Images}
Early approaches to lung tumor segmentation in PET/CT largely relied on traditional mathematical models and thresholding techniques\cite{erdi1997segmentation}. For example, analytic models such as the VMSBR (Volume/Motion/Source-to-Background Ratio) model~\cite{riegel2014defining} leveraged physical and geometric properties to adaptively estimate segmentation thresholds. These methods demonstrated improvements over static single-threshold techniques by better reflecting the motion envelope and imaging characteristics, especially for moving lung lesions.

However, such analytic frameworks are limited by handcrafted assumptions and parameterizations, which can struggle to generalize across diverse patient anatomies and complex imaging artifacts. With the emergence of deep learning, segmentation has shifted towards data-driven solutions capable of automatically capturing nuanced features and contextual information. The robust feature extraction and end-to-end learning capabilities that drive this advancement are primarily embodied in convolutional neural networks (CNNs), with fully convolutional networks (FCNs)\cite{kumar2019co, zhao2018tumor, xiang2022modality, li2020deep, zhong20183d, hooshangnejad2024exact} leading the way. In medical image segmentation, the predominant paradigm has become U-Net architectures, whose encoder-decoder framework utilizes lateral skip connections for integrating multi-scale feature representations and mitigating information loss throughout upsampling operations\cite{ronneberger2015u}. To address the locality constraints that convolutional operations impose on lung tumor segmentation, contemporary research has turned to Transformer-based methodologies, employing self-attention mechanisms for long-range spatial dependency modeling and global contextual information capture\cite{mao2024ms, bi2024co}. The recent emergence of Mamba has captured significant attention as a revolutionary state-space model that addresses the computational limitations of Transformers while maintaining their global modeling capabilities\cite{gu2023mamba}. Mamba employs selective state-space mechanisms (SSM) with linear computational complexity, enabling efficient processing of high-resolution medical images without the quadratic scaling issues inherent in self-attention mechanisms. This architecture demonstrates particular promise for lung tumor segmentation by combining the long-range dependency modeling capabilities of Transformers with the computational efficiency required for volumetric PET/CT analysis, while its selective scanning mechanism allows for adaptive feature selection based on input content relevance. Mamba-based models have been extensively deployed in medical image analysis, including SegMamba\cite{xing2024segmamba}, VM-UNet\cite{ruan2024vm}, U-Mamba\cite{ma2024u}, and Swin-UMamba\cite{liu2024swin}. Recent work utilizing CIPA has achieved state-of-the-art performance in PET/CT lung tumor GTV segmentation \cite{mei2025cross}.

\subsection{Transfer Learning}
Transfer learning has emerged as a pivotal strategy in medical image analysis, addressing the challenges of limited annotated data by leveraging knowledge from data-rich source domains to enhance performance in target domains with scarce annotations\cite{kora2022transfer, chen2019med3d}. In medical image analysis, this approach has demonstrated substantial benefits through pre-trained models from natural images or related medical tasks \cite{riaz2023lung, xie2018pre, jha2020doubleu}.

While early methods focused on ImageNet pre-training despite significant domain gaps\cite{jha2020doubleu}, recent research emphasizes medical-specific pre-training and cross-modal adaptation strategies\cite{guan2021domain}. Zoetmulder et al. demonstrated that optimal transfer learning for medical segmentation is achieved when the source domain and task closely resemble the target domain and task \cite{zoetmulder2022domain}. A critical application involves transferring knowledge between different radiotherapy target definitions, particularly GTV to IGTV segmentation. This transfer addresses the challenge that while GTV represents static tumor extent, IGTV encompasses respiratory motion patterns, requiring models to adapt from static to dynamic tumor characterization with limited IGTV training data.

\section{Dataset}

\subsection{Data Acquisition}
We utilized two PET/CT datasets in this study. The first dataset was obtained from the PCLT20K cohort, collected by the Department of Molecular Imaging and Medicine at a top-tier hospital. The examinations were performed between June 2016 and April 2020 using a GE Discovery Elite PET/CT scanner with fixed acquisition protocols. The second dataset is LUCID-PET/CT from the Johns Hopkins Hospital. Examinations for LUCID-PET/CT were conducted from June 2023 to December 2024 on multiple PET/CT scanners.

For the PCLT20K dataset, CT images had a fixed voxel size of 0.98 × 0.98 × 2.8 mm. PET images reconstructed using the ordered subset expectation maximization (OSEM) iterative algorithm had a fixed voxel size of 3.6 × 3.6 × 3.3 mm. In contrast, images in the LUCID-PET/CT dataset were acquired with varying voxel sizes due to the use of multiple scanners and heterogeneous acquisition protocols. Specifically, CT voxel sizes ranged from 0.98 × 0.98 × 3.00 mm to 1.52 × 1.52 × 5.00 mm, and PET voxel sizes had in-plane resolutions ranging from 1.65 × 1.65 mm to 5.47 × 5.47 mm with slice thicknesses between 2.79 mm and 5.00 mm. While a single PET reconstruction algorithm was employed for the PCLT20K dataset, a variety of reconstruction methods and parameter settings were used in the LUCID-PET/CT dataset, including OSEM, point spread function (PSF) with time-of-flight (TOF), Vue Point High Definition (VPHD), and others. Detailed scanner specifications and voxel size settings for both datasets are summarized in Table~\ref{tab:petct_scanner_voxel_short}. All patient-related privacy information, including patient names, physician names, and other identifiers, was fully de-identified during data collection in accordance with institutional privacy regulations.

The data annotation procedure consisted of three stages. For the PCLT20K dataset, the initial stage involved approximate delineation of the GTV, which was subsequently refined slice-by-slice on the PET/CT images during the second stage. In the third stage, the annotated volumes underwent an additional review to ensure consistency and accuracy. Similarly, annotation in the LUCID-PET/CT dataset was performed in three stages. First, each patient underwent a planning CT for radiotherapy; experienced clinical physicians then performed detailed, slice-by-slice delineation of the IGTV on these planning CT scans. PET/CT scans were first rigidly registered to the planning CT scans within the spine and lung regions, and then deformable registration was applied to the tumor region using the Velocity platform. After registration, an independent physician reviewed the registered PET/CT images. In cases where registration inconsistencies were identified, a third-stage physician repeated the IGTV delineation to ensure the highest possible accuracy and consensus in the annotations.

\begin{table*}[htbp]
\centering
\caption{Scanner specifications and voxel size settings for PCLT20K and LUCID-PET/CT datasets}
\label{tab:petct_scanner_voxel_short}
\resizebox{\textwidth}{!}{%
\begin{tabular}{llllcccccc}
\toprule
\textbf{Dataset} & \textbf{Manufacturer} & \textbf{Scanner Model} & \textbf{Modality} & \textbf{x$_{\min}$ (mm)} & \textbf{x$_{\max}$ (mm)} & \textbf{y$_{\min}$ (mm)} & \textbf{y$_{\max}$ (mm)} & \textbf{z$_{\min}$ (mm)} & \textbf{z$_{\max}$ (mm)} \\
\midrule
\multirow{2}{*}{PCLT20K} 
& GE MEDICAL SYSTEMS & Discovery Elite & CT  & 0.98 & 0.98 & 0.98 & 0.98 & 2.80 & 2.80 \\
& GE MEDICAL SYSTEMS & Discovery Elite & PET & 3.60 & 3.60 & 3.60 & 3.60 & 3.30 & 3.30 \\
\midrule
\multirow{6}{*}{LUCID-PET/CT} 
& GE MEDICAL SYSTEMS & Discovery Series & CT  & 0.98 & 1.37 & 0.98 & 1.37 & 3.75 & 3.75 \\
& GE MEDICAL SYSTEMS & Discovery Series & PET & 2.73 & 5.47 & 2.73 & 5.47 & 2.79 & 3.27 \\
& GE MEDICAL SYSTEMS & Optima Series & CT  & 1.37 & 1.37 & 1.37 & 1.37 & 5.00 & 5.00 \\
& GE MEDICAL SYSTEMS & Optima Series & PET & 3.65 & 3.65 & 3.65 & 3.65 & 3.27 & 3.27 \\
& SIEMENS & Biograph Series & CT  & 0.98 & 1.52 & 0.98 & 1.52 & 3.00 & 5.00 \\
& SIEMENS & Biograph Series & PET & 1.65 & 4.11 & 1.65 & 4.11 & 3.00 & 5.00 \\
\bottomrule
\end{tabular}%
}
\end{table*}

\subsection{Data Preprocessing}
In the PCLT20K dataset, CT images were preprocessed with a HU window of [-1200, 200] HU and PET images were scaled using SUV units, both normalized to [0, 255]. For the LUCID-PET/CT dataset, CT images used the standard lung window [-1350, 150] HU, while PET images were normalized using ROI-specific maximum SUV values to mitigate the influence of high-uptake regions like the heart. Empirical quality control criteria were implemented to exclude IGTV ROIs with volumes $< 3.0$ cc or maximum SUV $< 3.0$ to ensure clinically relevant lesions. For each valid ROI, 3D binary masks were generated from DICOM contour sequences and co-registered with corresponding CT and PET slices based on spatial coordinates. Tumor volumes were calculated from the 3D binary masks, accounting for anisotropic voxel dimensions. All images were saved in standardized 512x512 PNG format, accompanied by comprehensive metadata that included ROI statistics and volumetric measurements for subsequent analysis.

\subsection{Data Statistics}

The LUCID-PET/CT dataset comprised 60 patients with lung tumors, yielding 60 IGTV regions of interest after quality control filtering. The dataset contained a total of 1,067 axial slices, with an average of $17.8 \pm 8.2$ slices per ROI (range: 9--47 slices). Tumor volumes ranged from 3.60 to 822.86~cc with a median volume of 11.81~cc, and maximum SUV values ranged from 3.12 to 23.98 (mean: $8.25 \pm 4.3$). All ROIs met the inclusion criteria of volume $\geq 3.0$~cc and maximum SUV $\geq 3.0$ to ensure clinical relevance.

\begin{figure}[htbp]
\centering
\includegraphics[width=\columnwidth]{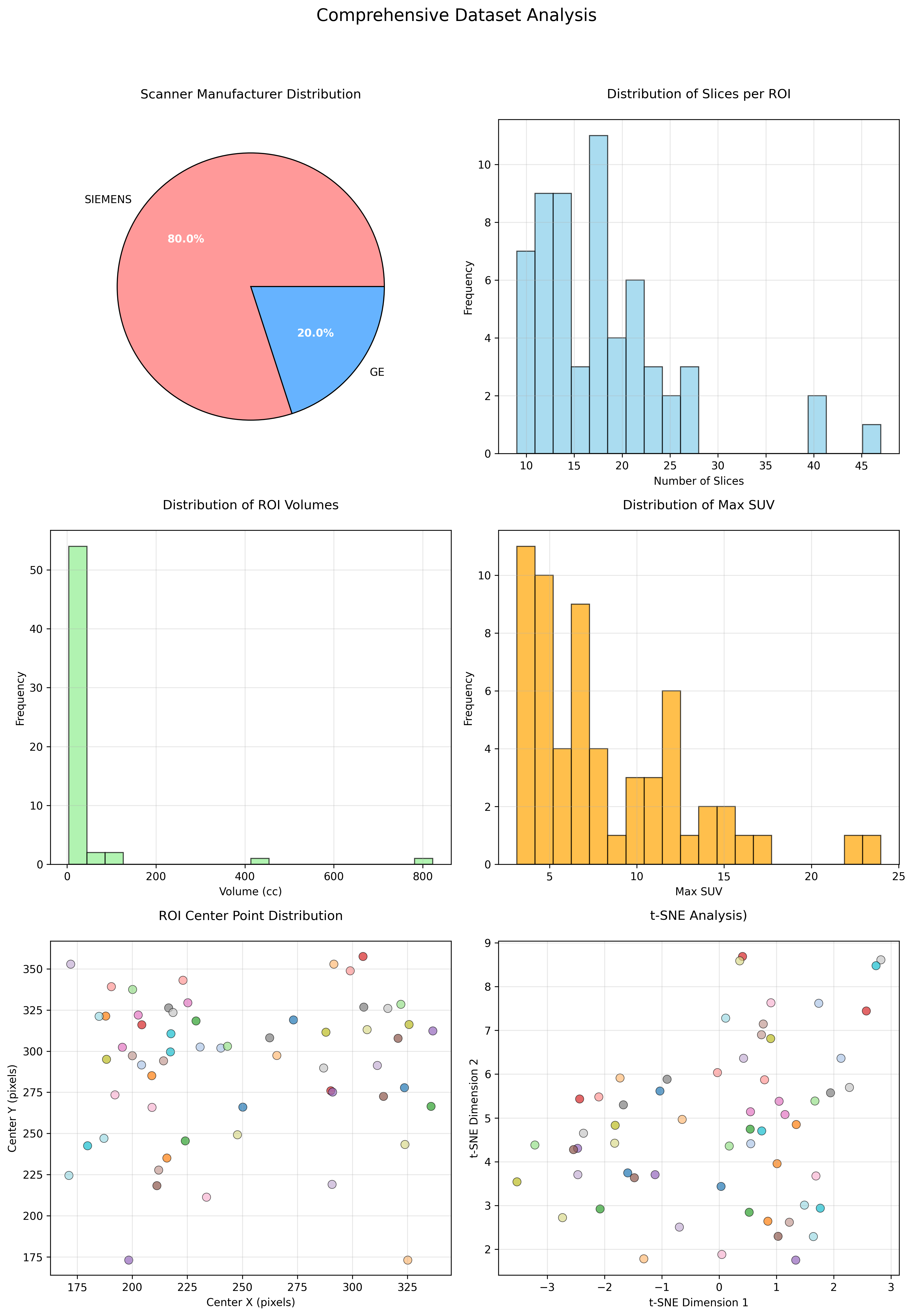}
\caption{Comprehensive statistical analysis of the LUCID-PET/CT dataset. (Top left) Scanner manufacturer distribution showing the proportion of GE Medical Systems and SIEMENS scanners used across the dataset. (Top right) Distribution of the number of slices per ROI, indicating the variability in tumor coverage requirements. (Middle left) Tumor volume distribution demonstrating the right-skewed nature typical of clinical datasets. (Middle right) Maximum SUV distribution reflecting the heterogeneous metabolic activity of lung tumors. (Bottom left) ROI center point distribution across patients, illustrating the spatial diversity of tumor locations within the lung field. (Bottom right) t-SNE visualization of dataset diversity using 11 features, demonstrating distinct clustering patterns that validate the multi-dimensional heterogeneity of the dataset.}
\label{fig:jhu_dataset_statistics}
\end{figure}

Figure~\ref{fig:jhu_dataset_statistics} illustrates the comprehensive dataset characteristics. The scanner manufacturer distribution shows that GE Medical Systems and SIEMENS scanners were used for 80\% and 20\% of cases, respectively, demonstrating the multi-vendor nature of the dataset. The distributions of tumor volumes and SUV values exhibit right-skewed patterns typical of clinical datasets, with most lesions being relatively small ($<50$~cc) and most tumors showing SUV values between 3--10, while some cases demonstrated exceptionally large volumes ($>800$~cc) or high metabolic activity (SUV $>20$), reflecting the heterogeneous nature of lung tumors in clinical practice.

The spatial analysis reveals that ROI center points are distributed across different anatomical locations within the lung field, with coordinates ranging from [X-range] to [Y-range] pixels, indicating good spatial diversity across the dataset. The t-SNE visualization of the 11-dimensional feature space demonstrates clear clustering patterns that reflect the underlying heterogeneity of the dataset. Different patients exhibit distinct feature distributions in the reduced-dimensional space, validating the dataset's diversity in terms of tumor characteristics, spatial distribution, and metabolic properties. This multi-dimensional diversity ensures that our model is trained on a representative sample of clinical IGTV segmentation scenarios, enhancing its generalizability to real-world applications.

\section{Method}

\subsection{Overall Framework}

Our approach addresses the challenges of IGTV segmentation through a comprehensive framework consisting of three key components: (1) a transfer learning strategy that leverages knowledge from large-scale GTV datasets to overcome limited IGTV annotations, (2) a 2.5D segmentation paradigm that processes multiple consecutive slices to capture spatial context across the slice dimension, and (3) a lightweight SIM that models inter-slice relationships to improve boundary delineation in regions with ambiguous PET signal intensity.

\begin{figure*}[t]
    \centering
    \includegraphics[width=1\textwidth]{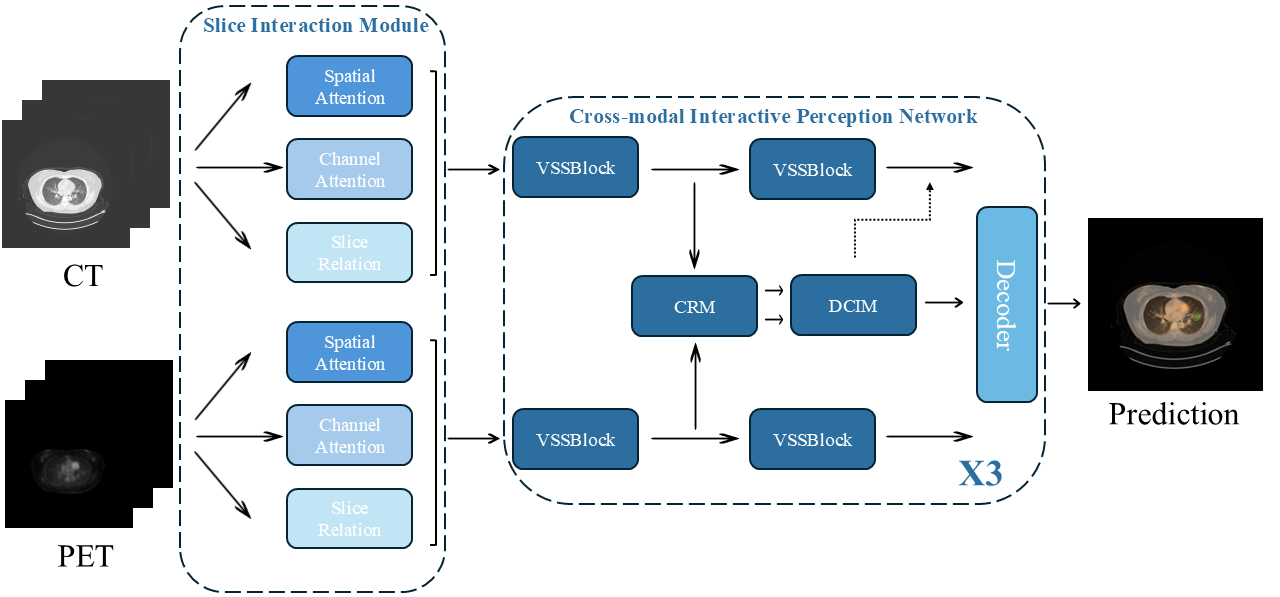}
    \caption{Overview of the proposed IGTV segmentation framework, which employs a 2.5D segmentation paradigm and incorporates a lightweight SIM to model inter-slice relationships and improve boundary delineation.}
    \label{fig:overall_framework}
\end{figure*}

The overall model architecture of our IGTV segmentation framework is depicted in Figure~\ref{fig:overall_framework}. The model employs a 2.5D segmentation paradigm, in which three consecutive slices are jointly processed to capture spatial context and anatomical continuity across the slice dimension. To further enhance segmentation performance, especially in regions with ambiguous PET signal intensity, we introduce SIM, specifically designed to model inter-slice relationships and facilitate robust boundary delineation by aggregating information across adjacent slices. This integrated design enables the network to leverage both intra-slice and inter-slice features for more accurate IGTV segmentation.

\subsection{Base Network Architecture}

We build upon the CIPA architecture \cite{mei2025cross} as our foundation due to its demonstrated state-of-the-art performance in PET/CT lung tumor segmentation. CIPA employs a dual-branch encoder design with MAMBA backbones for processing PET and CT modalities separately, followed by CRM and DCIM for effective fusion of metabolic and anatomical information.

The MAMBA architecture provides several advantages for medical image segmentation: (1) linear computational complexity $O(L)$ compared to the quadratic scaling $O(L^2)$ of self-attention mechanisms in Transformers, (2) selective state-space mechanisms that enable adaptive feature selection based on input content, and (3) efficient processing of high-resolution medical images while maintaining global context modeling capabilities through its recurrent structure.

\subsection{2.5D Segmentation with Slice Interaction Module}

\subsubsection{2.5D Input Configuration}
Unlike conventional 2D approaches that process single slices independently, our method adopts a 2.5D strategy that takes three consecutive slices as input: $\{\mathbf{I}_{t-1}, \mathbf{I}_t, \mathbf{I}_{t+1}\}$, where $\mathbf{I}_t$ represents the target slice for segmentation and $\mathbf{I}_{t-1}$, $\mathbf{I}_{t+1}$ are the adjacent slices. For each modality (PET and CT), these three slices are concatenated along the channel dimension to form the input tensor $\mathbf{X} \in \mathbb{R}^{B \times 3 \times H \times W}$, providing richer spatial context that is essential for accurate IGTV delineation accounting for respiratory motion patterns.

\subsubsection{Slice Interaction Module Design}
To effectively model inter-slice relationships and enhance boundary delineation in regions with weak PET signal intensity, we propose a lightweight SIM integrated into the encoder. The SIM consists of three complementary branches that capture different aspects of slice interactions:

\textbf{Channel Attention Branch:} This branch adaptively weights the importance of each slice through a channel-wise attention mechanism:
\begin{align}
\mathbf{A}_{ch} &= \sigma(\mathbf{W}_2 \cdot \text{ReLU}(\mathbf{W}_1 \cdot \text{GAP}(\mathbf{X}))) \\
\mathbf{X}_{ca} &= \mathbf{X} \odot \mathbf{A}_{ch}
\end{align}
where $\text{GAP}(\cdot)$ denotes global average pooling, $\mathbf{W}_1 \in \mathbb{R}^{C/r \times C}$ and $\mathbf{W}_2 \in \mathbb{R}^{C \times C/r}$ are learnable parameters with reduction ratio $r=2$, $\sigma(\cdot)$ is the sigmoid activation function, and $\odot$ represents element-wise multiplication.

\textbf{Spatial Attention Branch:} This branch identifies salient spatial regions across slices by learning spatial relationships:
\begin{align}
\mathbf{A}_{sp} &= \sigma(\text{BN}(\text{Conv}_{7 \times 7}(\mathbf{X}))) \\
\mathbf{X}_{sa} &= \mathbf{X} \odot \mathbf{A}_{sp}
\end{align}
where $\text{Conv}_{7 \times 7}(\cdot)$ represents a $7 \times 7$ convolution that reduces the channel dimension to 1, and $\text{BN}(\cdot)$ denotes batch normalization.

\textbf{Slice Relation Branch:} This branch employs depthwise convolutions to model complex inter-slice dependencies while maintaining computational efficiency:
\begin{equation}
\mathbf{X}_{rel} = \text{DWConv}_{3 \times 3}(\text{ReLU}(\text{BN}(\text{DWConv}_{1 \times 1}(\mathbf{X}))))
\end{equation}
where $\text{DWConv}_{k \times k}(\cdot)$ represents depthwise convolution with kernel size $k$, processing each slice channel independently while preserving spatial relationships.

\textbf{Feature Fusion:} The outputs from all three branches are combined with the original input through a weighted residual connection:
\begin{equation}
\mathbf{X}_{SIM} = \mathbf{X} + \alpha \cdot \mathbf{X}_{ca} + \beta \cdot \mathbf{X}_{sa} + \gamma \cdot \mathbf{X}_{rel}
\end{equation}
where $\alpha = 0.3$, $\beta = 0.3$, and $\gamma = 0.4$ are empirically determined weights that balance the contributions from different attention mechanisms. The residual connection preserves original information while the weighted combination allows adaptive emphasis on different types of inter-slice relationships.

\subsection{Transfer Learning Strategy}

To address the scarcity of annotated IGTV data while leveraging abundant GTV annotations, we implement a two-stage transfer learning approach that enables effective knowledge transfer from GTV to IGTV segmentation tasks.

\subsubsection{Stage 1: Pre-training on GTV Dataset}
The base CIPA network is initially trained on the large-scale PCLT20K dataset containing 21,930 PET/CT image pairs with GTV annotations using standard 2D input processing. During this stage, the network learns robust multimodal feature representations for lung tumor segmentation. The training objective combines Dice loss and binary cross-entropy loss:
\begin{equation}
\mathcal{L}_{GTV} = \mathcal{L}_{Dice}(\mathbf{P}_{GTV}, \mathbf{Y}_{GTV}) + \mathcal{L}_{BCE}(\mathbf{P}_{GTV}, \mathbf{Y}_{GTV})
\end{equation}
where $\mathbf{P}_{GTV}$ and $\mathbf{Y}_{GTV}$ represent predicted and ground truth GTV masks, respectively.

\subsubsection{Stage 2: Fine-tuning on IGTV Dataset}
The pre-trained network is adapted for IGTV segmentation using our LUCID-PET/CT dataset containing 1,067 image pairs from 60 patients. The dataset is partitioned into 70\% training, 15\% validation, and 15\% testing splits at the patient level using a fixed random seed for reproducibility.

During transfer learning, we extend the 2D pre-trained model to the 2.5D architecture by integrating the proposed SIM and modifying the input processing to handle three consecutive slices. All compatible network parameters are initialized from the pre-trained GTV model, while newly added 2.5D components are randomly initialized. The entire network is then fine-tuned end-to-end using the same loss function:
\begin{equation}
\mathcal{L}_{IGTV} = \mathcal{L}_{Dice}(\mathbf{P}_{IGTV}, \mathbf{Y}_{IGTV}) + \mathcal{L}_{BCE}(\mathbf{P}_{IGTV}, \mathbf{Y}_{IGTV})
\end{equation}

A significantly reduced learning rate is employed during fine-tuning to preserve learned multimodal representations while adapting to IGTV-specific characteristics. Model selection is performed based on validation IoU, with the best-performing model evaluated on the held-out test set.

\subsection{Implementation Details}

\subsubsection{Training Configuration}
The network is implemented using PyTorch and trained on NVIDIA A100 GPUs with Automatic Mixed Precision enabled for computational efficiency. 

For pre-training on the PCLT20K dataset, we employ: initial learning rate of $1 \times 10^{-3}$ with cosine annealing scheduler, AdamW optimizer with $\beta_1=0.9$, $\beta_2=0.999$, weight decay of $1 \times 10^{-2}$, and batch size of 16.

For fine-tuning on the IGTV dataset, the learning rate is reduced to $6 \times 10^{-5}$ to preserve pre-trained features while adapting to IGTV characteristics, with batch size of 4 due to increased memory requirements of the 2.5D approach. The 2.5D input configuration processes three consecutive slices with a stack depth of 3, resulting in 6 input channels (2 modalities $\times$ 3 slices). Training is conducted for 30 epochs with early stopping based on validation IoU.

\subsubsection{Augmentation}
Data augmentation is applied consistently during both pre-training and fine-tuning phases to enhance model generalization. The augmentation pipeline includes: (1) random affine transformations with spatial shift within $\pm 10\%$ of image dimensions and isotropic scaling factors sampled from [0.9, 1.1], (2) random horizontal flipping with 50\% probability, and (3) random cropping that extracts 70\%-90\% of the original image area followed by bilinear interpolation to restore original dimensions. Each augmentation operation is applied with 50\% probability to maintain training stability. 

\section{Results}

\subsection{Experimental Setup}

\begin{figure*}[!htbp]
\centering
\includegraphics[width=1\textwidth]{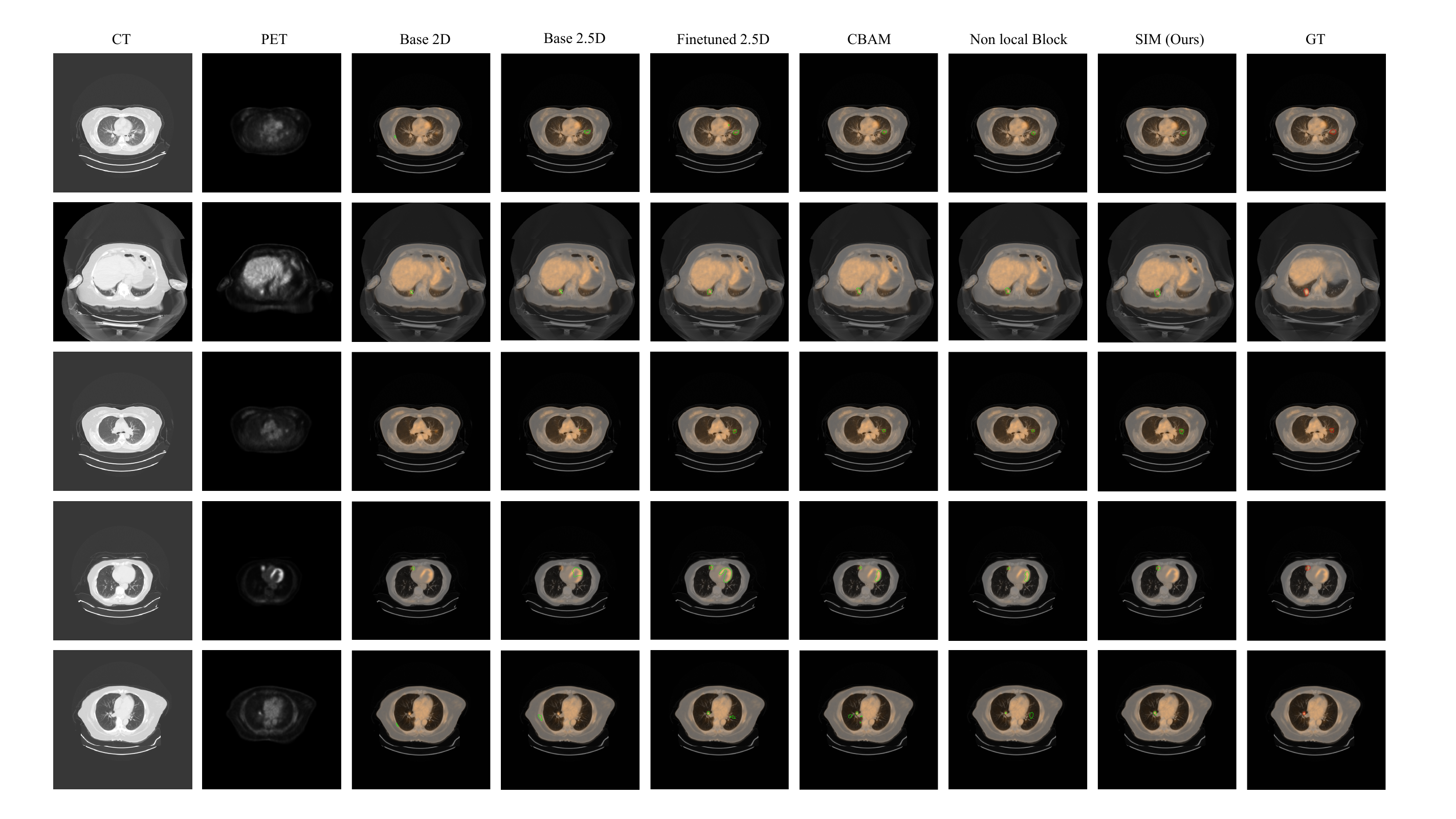}
\caption{Visual comparison of segmentation results from different models on representative PET/CT slices from the LUCID-PET/CT test set. Ground truth contours are shown in red, predicted segmentation in green. From left to right: input CT, input PET, ground truth, Baseline 2D, Baseline 2.5D, Finetuned 2.5D, Finetuned 2.5D + CBAM, Finetuned 2.5D + Non-Local Block, Finetuned 2.5D + SIM (Ours).}
\label{fig:seg-visual}
\end{figure*}

All pre-training experiments are conducted on the CIPA dataset, while fine-tuning is performed on the LUCID-PET/CT dataset containing 1,067 PET/CT image pairs from 60 patients.

We evaluate our approach using standard segmentation metrics: Intersection over Union (IoU), Dice, pixel accuracy (Acc), and 95th percentile Hausdorff Distance (HD95). Specifically, the metrics are defined as follows:
\begin{align}
\text{IoU} &= \frac{|P \cap G|}{|P \cup G|} \\
\text{Dice} &= \frac{2|P \cap G|}{|P| + |G|} \\
\text{Acc} &= \frac{\sum_{i} \mathbb{I}(p_i = g_i)}{N} \\
\text{HD95}(P, G) &= \max\{h_{95}(P, G), h_{95}(G, P)\}
\end{align}

Here, $P$ denotes the set of predicted pixels, $G$ denotes the set of ground truth pixels, $\mathbb{I}(\cdot)$ is the indicator function, $p_i$ and $g_i$ represent the predicted and ground truth labels for the $i$-th pixel respectively, $N$ is the total number of pixels, and $\|\cdot\|_2$ represents the Euclidean distance..

\subsection{Ablation Study}

We conduct a comprehensive ablation study to evaluate the effectiveness of each component in our proposed framework. The results are presented in Table~\ref{tab:ablation}, where the "Finetune" column indicates whether model fine-tuning is applied (\checkmark) or not ($\times$).

\begin{table*}[htbp]
\centering
\caption{Ablation study results on the LUCID-PET/CT test set, evaluating the impact of fine-tuning and different attention modules (CBAM, Non-Local Block, SIM). All models use 2.5D input with three consecutive slices unless otherwise specified. Finetune (\checkmark) indicates fine-tuning is applied. The best scores are in bold.}
\label{tab:ablation}
\begin{tabular}{lccccc}
\hline
\textbf{Method} & \textbf{Finetune} & \textbf{IoU} & \textbf{Dice} & \textbf{Acc} & \textbf{HD95} \\
\hline
Baseline 2D & $\times$ & 0.238 & 0.385 & 0.646 & 36.40 \\
Baseline 2.5D & $\times$ & 0.240 & 0.387 & 0.686 & 31.39 \\
Baseline 2.5D & \checkmark & 0.363 & 0.532 & 0.823 & \textbf{22.21} \\
Baseline 2.5D + CBAM & \checkmark & 0.296 & 0.457 & 0.787 & 48.68 \\
Baseline 2.5D + Non Local Block & \checkmark & 0.369 & 0.539 & 0.772 & 25.93 \\
Baseline 2.5D + SIM (\textbf{Ours}) & \checkmark & \textbf{0.438} & \textbf{0.609} & \textbf{0.846} & 24.81 \\

\hline
\end{tabular}
\end{table*}

For models without fine-tuning, the baseline 2D and 2.5D configurations achieve similar performance, with IoU scores of 0.238/0.240 and Dice of 0.385/0.387, respectively. While the improvement from 2D to 2.5D is modest in the absence of fine-tuning, the slight advantage suggests that incorporating additional spatial context can still provide incremental benefits.

In contrast, models with fine-tuning exhibit substantially better results. The finetuned 2.5D baseline achieves an IoU of 0.363 and a Dice of 0.532, demonstrating the importance of both spatial context and fine-tuning for segmentation performance. The incorporation of our SIM further improves results, achieving the highest scores overall: IoU of 0.438, Dice of 0.609, and pixel accuracy of 0.846. Compared to the finetuned 2.5D baseline, this represents relative improvements of 20.7\% in IoU and 14.5\% in Dice, confirming the effectiveness of our attention mechanism for capturing inter-slice dependencies.

We also compare our SIM module against alternative attention mechanisms under finetuned settings. CBAM achieves inferior performance compared to the 2.5D baseline, with IoU of 0.296, Dice of 0.457, and a higher HD95 value of 48.68. The Non-Local Block improves performance over CBAM (IoU of 0.369, Dice of 0.539), but is still lower than our SIM module. These results demonstrate that our specifically designed slice-aware interaction mechanism is more effective for PET/CT lesion segmentation than generic attention modules, as it explicitly models the relationships between adjacent slices in the 2.5D context.

\subsection{Qualitative Comparison}
Figure~\ref{fig:seg-visual} presents visual comparisons of segmentation results produced by different models. These examples are selected from the LUCID-PET/CT test set and demonstrate the influence of each module and training strategy on lesion delineation. The baseline 2D and 2.5D models without fine-tuning are easily affected by non-tumor signals and often fail to capture lesion boundaries accurately. In contrast, the finetuned models, especially the one incorporating the proposed SIM, achieve more precise and complete segmentation results. The selected samples include marginal slices, which typically have low PET signals, as well as slices that contain prominent organ signals that can mislead the models. Our method demonstrates robust performance under these challenging conditions and accurately delineates lesions.

\section{Discussion}

Our multimodal slice interaction network with transfer learning shows strong performance for IGTV segmentation in PET/CT, but several limitations remain.

First, the fine-tuning step is limited by the relatively small size and diversity of the available IGTV dataset. This may affect generalization to rare or atypical cases, and more diverse annotated data would likely improve robustness in real-world applications.

Second, our fixed 2.5D input with three consecutive slices may miss long-range anatomical information or relationships spanning more than three slices. Further work could explore dynamic slice selection or full 3D architectures to address this.

\begin{figure}[htbp]
    \centering
    \includegraphics[width=1\columnwidth]{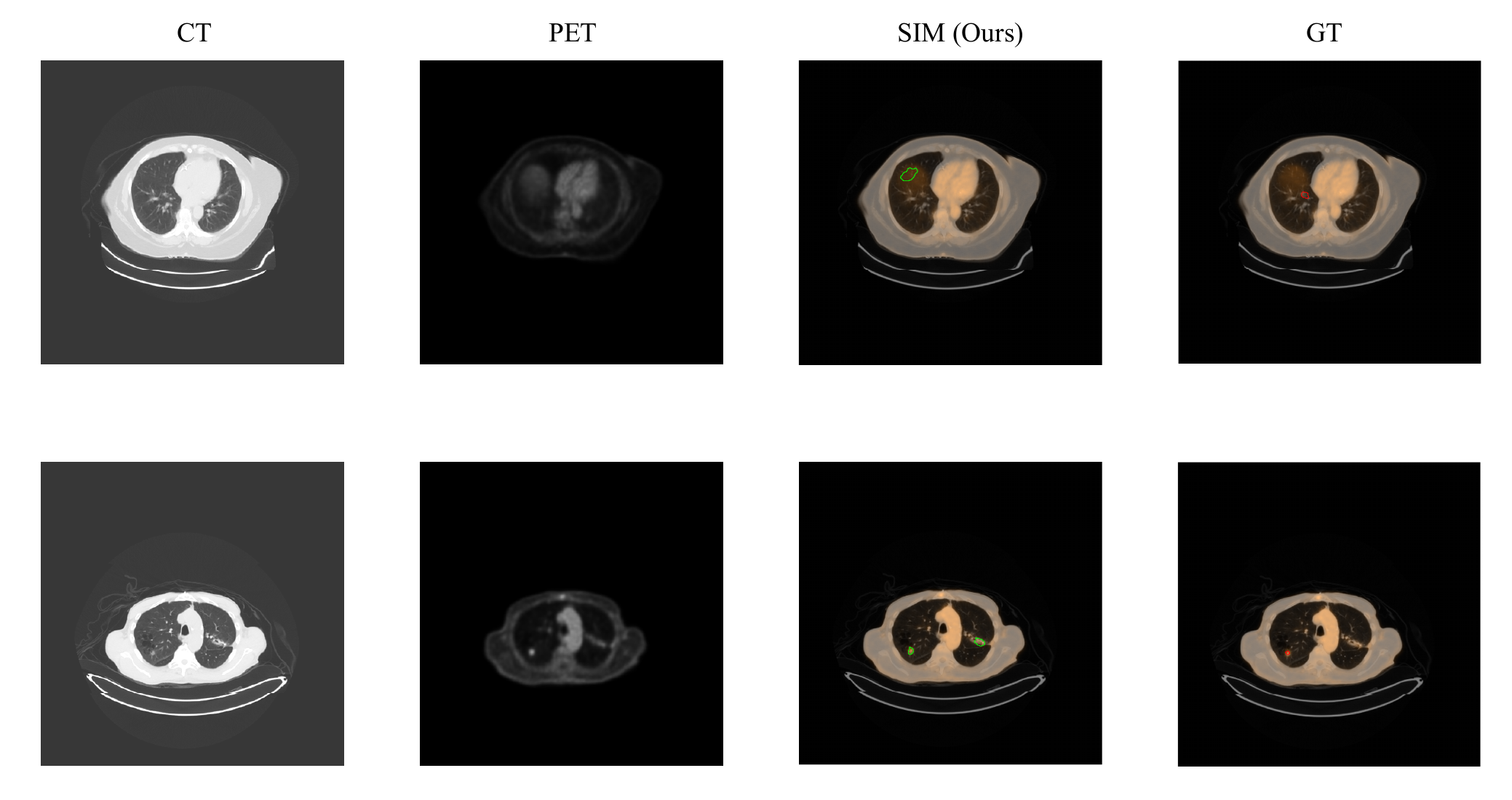}
    \caption{Representative failure cases. Left to right: input CT, input PET, model prediction, ground truth. The model misses tumor boundaries and includes false positives when PET signals are weak and adjacent organ uptake is high.}
    \label{fig:case-fail}
\end{figure}

The model also still struggles in highly challenging cases, such as those with low PET signal at the tumor boundary or with strong adjacent organ uptake. Figure~\ref{fig:case-fail} shows two representative failure cases: the model misses part of the lesion or produces false positives due to weak peripheral PET signals and interference from nearby organs. These results highlight areas for future improvement in multi-modal fusion and contextual modeling.

Finally, while our model shows promising segmentation accuracy, further clinical validation is needed to assess its practical impact in real radiotherapy planning.

\section{Conclusion}

In this work, we proposed a multimodal slice interaction network enhanced by transfer learning for precise segmentation of IGTV in lung cancer PET/CT imaging. Leveraging large-scale annotated datasets for pre-training and a private IGTV cohort for fine-tuning, our framework effectively overcomes data scarcity and adapts to the challenging characteristics of IGTV delineation. The introduction of SIM within a 2.5D segmentation paradigm further improves model performance, especially for slices with weak PET signal or complex background.

Comprehensive experiments demonstrate that our approach achieves superior segmentation accuracy compared to conventional baselines and alternative attention mechanisms. The study is still constrained by factors such as an insufficient IGTV dataset size and the inadequate modeling of long-range spatial relationships. Future work will focus on expanding the dataset, integrating more advanced 3D spatial strategies, and performing extensive evaluation of clinical utility for radiotherapy planning.

Overall, our approach provides a promising direction for advancing automated IGTV segmentation in lung cancer, with potential applications in other multimodal imaging and challenging clinical scenarios.

\section*{Acknowledgment}
None.

\bibliographystyle{IEEEtran}

% \bibliography{ref}
\end{document}